# Deep Cost-sensitive Learning for Wheat Frost Detection

Shujian Cao
School of Computer Science and Information Engineering
Hefei University of Technology
Hefei, China
shujian_cao@mail.hfut.edu.cn

Lin Cui
School of Information Engineering
Suzhou University
Suzhou, China
cl@ahszu.edu.cn

Haipeng Liu*
School of Computer Science and Information Engineering
Hefei University of Technology
Hefei, China
hpliu_hfut@hotmail.com

*Abstract*—Frost damage is one of the main factors leading to wheat yield reduction. Therefore, the detection of wheat frost accurately and efficiently is beneficial for growers to take corresponding measures in time to reduce economic loss. To detect the wheat frost, in this paper we create a hyperspectral wheat frost data set by collecting the data characterized by temperature, wheat yield, and hyperspectral information provided by the handheld hyperspectral spectrometer. However, due to the imbalance of data, that is, the number of healthy samples is much higher than the number of frost damage samples, a deep learning algorithm tends to predict biasedly towards the healthy samples resulting in model overfitting of the healthy samples. Therefore, we propose a method based on deep cost-sensitive learning, which uses a one-dimensional convolutional neural network as the basic framework and incorporates cost-sensitive learning with fixed factors and adjustment factors into the loss function to train the network. Meanwhile, the accuracy and $F_1$ score are used as evaluation metrics. Experimental results show that the detection accuracy and the $F_1$ score reached 0.943 and 0.623 respectively, this demonstration shows that this method not only ensures the overall accuracy but also effectively improves the detection rate of frost samples.

## I. INTRODUCTION

Frosts during the growth of wheat can lead to significant yield reductions, so accurately and efficiently detecting wheat frost and the adoption of measures such as timely harvesting, fodder storage, or grazing can reduce economic losses to growers. In wheat frost research, existing methods for detecting wheat frost rely primarily on human judgement based on crop characteristics and experience, but this method is affected by factors such as crop growth stage, personal experience and weather conditions that make it difficult to achieve ideal identification efficiency and accuracy [1]. Especially in modern large-scale agriculture, the traditional method of identifying crop diseases based on physical vision requires a lot of humans, material, and financial resources, and it is difficult to be fully extended.

Modern agriculture can use hyperspectral remote sensing technology to assess whether crops are being affected by such factors as nitrogen [2], water [3] salt [4] and heat [5]. On this basis, for the task of wheat frost detection, we construct a hyperspectral wheat frost dataset by extracting, pre-processing, and creating labels from hyperspectral remote sensing data of wheat collected using a handheld hyperspectrometer on a trial field in Western Australia, in which each sample contains 2151 dimensions of spectral information, which can accurately reflect the frost condition of wheat. Currently, there are two main ways to process hyperspectral data: traditional machine learning methods and deep learning techniques. Due to the high dimensionality and large volume of hyperspectral remote sensing data, it is inefficient to use traditional machine learning methods to process hyperspectral data. In recent years, both the theory and techniques of deep learning have made rapid progress, and as it can automatically extract deep features in hyperspectral data, this has led to significant results in tasks such as classification and detection of hyperspectral images [6]. Therefore, we employ a deep convolutional neural network for the detection of frost samples in hyperspectral data. Specifically, the 2151 dimensional spectral information of wheat is used as the input of the convolutional neural network, and then its depth features are extracted through multiple convolutions, activation, and pooling operations, and the final depth features are then passed through a layer of fully connected layers and softmax layers to obtain the detection results.

However there is also a sample imbalance in the wheat hyperspectral data in that there are far more samples of healthy wheat than of frosty wheat. This allows the model to focus overly heavily on the majority of samples during training, a bias that may result in the model being unable to predict minority class of samples, whereas in wheat frost detection task, frosty samples have more important information, meaning growers cannot act on the results if frosty samples are not effectively identified [7]. In the face of this imbalance, the current research focuses on two aspects: data level, which aims at adjusting data distribution, and algorithm level, which aims at optimizing target functions to improve the performance of deep learning models in detection tasks.

* Haipeng Liu is the corresponding author

The main method of adjusting the distribution of data at the data level is sample sampling, i.e., reducing the majority or increasing minority class samples to obtain a relatively balanced dataset. However, the effect of resampling at the data level to improve the detection accuracy is limited, and existing research is more and more inclined to solve class imbalance problem at the algorithm level. One of the mainstream techniques to address class imbalance problems at the algorithmic level is cost-sensitive learning [8]. The ultimate goal is not to minimize training loss, but to minimize the overall misclassification cost during training by increasing the misclassification costs for a small number of samples or reducing the misclassification costs for a large number of samples during training, thereby increasing the generalization ability of the model [9].

Therefore, a cost-sensitive learning algorithm on deep learning is proposed to solve the problem of sample imbalance problem. Instead of general cost-sensitive learning that directly sets the cost of various misclassifications, we balance this by introducing a fixed factor to increase the misclassification cost of a small number of frost samples and an adjustment factor to adjust the misclassification cost of a small number of samples in real time. These two factors enable the algorithm to differentiate between healthy and frosty samples effectively.

We constructed a hyperspectral wheat frost dataset and applied a deep cost-sensitive learning approach to address the problems of high data dimensionality and unbalanced samples in the data, using a one-dimensional convolutional neural network as the underlying framework to alleviate the unbalanced distribution of labels in the frost dataset by introducing cost coefficients into the loss function. In addition, to verify the effectiveness of the method, four sets of experiments were designed in this paper and a combination of accuracy and $F_1$ scores were used as evaluation metrics for the performance of the convolutional neural network. The experimental results show that the method in this paper achieves a correct rate and $F_1$ score of 0.943 and 0.623 respectively, and is an effective detection method to identify healthy samples and frosty samples.

## II. DATASETS AND METHODS

### A. Datasets

The raw data used in this paper was provided by the Australian Agricultural Research Federation [10], and we constructed a hyperspectral wheat frost dataset by data collation and processing of the raw data.

The experiment to generated the raw data was carried out on a farm in Western Australia, as shown in Fig. 1, where each small rectangle represents a small growing area and each area is planted with one type of wheat, and several small areas are combined to form a large area labeled A, B, C, D, E, F, G, H. A total of 16 commercial wheat varieties were used in the trial, all of which were planted in areas C, D, E, and F. The wheat varieties in each of the four areas were randomly distributed.

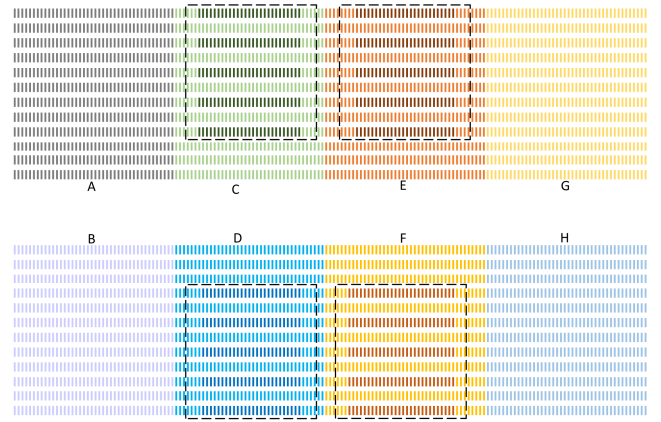

Figure 1. The location of plots in the field

The raw data consisted of 2151 dimensional data collected by hyperspectral data acquisition using an ASD FieldSpec® 4 Hi-Res hyperspectrometer, the temperature of each plot at different moments in time, and the corresponding yield of the plot.

We use a combination of temperature and crop yield as indicators to label the data set. A plot is considered to have suffered a frost when the temperature is below zero for a certain period, but the mere fact that it is below zero for a certain period or a short period does not affect the crops in the plot. There may also be cases where the temperature is below freezing but the crop is not frosty, so after the frost has occurred, the yield is monitored for the four areas sampled C, D, E, and F The total number of measurements for the 16 wheat species was 1000, with 940 healthy samples and 60 frosty samples.

### B. Model construction

For each sample in the hyperspectral wheat frost dataset, we use 2151 band values as its features, and the local features of these samples are correlated with each other. RNN (Recurrent Neural Network) [11] or CNN (Convolutional Neural Network) [12] can handle such samples with the correlation between features very well. However, due to the time structure of the RNN, it is difficult to obtain effective computational power in the face of high-dimensional samples in hyperspectral wheat frost dataset [13]. Instead, CNN reduces the number of network parameters that need to be trained through weight sharing, thus reducing the consumption of computational resources. Therefore, we adopt the CNN approach to feature extraction and, and since hyperspectral wheat samples require only one-dimensional feature extraction, the basic framework used is 1D-CNN (One Dimensional Convolution Neural Network) [14] and the sequences are converted to a 2151×1 format. In addition, the pooling method used in this paper is max-pooling because of the need to maximize the retention of feature-to-feature differences.

The 1D-CNN framework used in this paper is shown in Fig. 2. The network consists of three convolutional layers, three pooling layers, and one fully connected layer. The first convolutional layer has a dimension of n × 2151 × 1, n being the number of input samples, the size of the convolutional operator in the first layer is 7 × 1, the number of operators is 64 and the activation function is ReLU. The output of the first layer is then downsampled as the input of the pooling layer, and the size of the pooling operator in the pooling layer is 9 × 1. The number of convolutional kernels in the second and third layers is adjusted to 128 and 32, respectively, and the number of pooling kernels in the second and third layers is adjusted to 128 The second and third layers have the same structure as the first layer except that the number of convolutional kernels is adjusted to 128 and 32, respectively, and the pooling parameters are adjusted to 5 × 1 and 7 × 1. In addition, to maintain the same distribution of inputs in each layer during training, a BatchNorm layer is introduced after each convolutional layer. A fully-connected layer plus a Softmax layer is connected after the third layer to regress the high-dimensional features to the corresponding class probability values.

C. *Cost-sensitive loss functions*

In the hyperspectral wheat frost dataset, the number of frosty samples marked with 1 is 60, while the number of healthy samples marked with 0 is 940, with a ratio of 1:16 between the two types of samples, which also means that the occurrence of healthy samples is higher than that of frosty samples. As a result, the model tends to be biased towards the majority of samples during training, resulting in poor generalization of the model. To address this problem, we introduce cost-sensitive learning, i.e. by introducing cost coefficients that make the misclassification costs of different types of samples have large differences.

1) *Cost matrix*

The cost matrix [16] for the binary anomaly detection of healthy and frost samples of wheat in this paper is shown in TABLE I, where four classification costs are involved, namely $C_{00}$, $C_{01}$, $C_{10}$, and $C_{11}$ ($C_{ij}$ denotes the cost arising from classifying class i into class j), e.g. $C_{00}$ denotes the cost of misclassifying a healthy sample into an impaired sample. It can be considered that $C_{00}$ and $C_{11}$ do not incur a misclassification cost, i.e. $C_{00} = C_{11} = 0$, and the misclassification cost for impaired samples is greater than the misclassification cost for normal samples, i.e. $C_{10} \geq C_{01}$.

TABLE I. COST MATRIX

| Actual category | Forecast Type | |
|---|---|---|
| | Healthy (0) | Frosted (1) |
| Healthy (0) | $C_{00}$ | $C_{01}$ |
| Frosted (1) | $C_{10}$ | $C_{11}$ |

The loss of classifying x into i in cost-sensitive learning can be expressed as

$$L(x, i) = P(j|x) C_{ij} \quad (1)$$

where $(x, i)$ denotes the division of $x$ is divided into $i$ category. $P(j|x)$ indicates that $x$ is the probability of belonging to $j$ of the posterior probability ($j \neq i$), the smaller the value of $P(j|x)$, the larger the value of $P(i|x)$. $C_{ij}$ indicates that it will $i$ the wrong score $j$ of the actual loss. In equation (1), the value of the bias factor $C_i$ is added so that the goal of the network is not only how to obtain the $P(j|x)$ the minimum value of the prediction, but also to consider the loss due to incorrect prediction results $C_{ij}$.

2) *Cost loss functions*

We introduce a cost factor $W_i$ on top of the standard cross-entropy loss function [15] to make the misclassification of frosty samples more costly. The cost sensitive binary loss (CSBL) function designed based on cross-entropy is

$$L_{log} = -\frac{1}{N} \sum_{i=1}^{N} y_i \log \hat{y}_i + (1 - y_i) \log(1 - \hat{y}_i) \quad (2)$$

$$L_w = -\frac{1}{N} \sum_{i=1}^{N} W_i (y_i \log \hat{y}_i + (1 - y_i) \log(1 - \hat{y}_i)) \quad (3)$$

where $L_{log}$ denotes the standard cross-entropy loss function. $L_w$ denotes the objective function value. $W$ is the weight to be trained in the one-dimensional convolutional neural network. $N$ denotes the number of samples. $\hat{y}_i$ denotes the probability that the first $i$ the probability that the first sample is predicted to be a frosty sample. $y_i$ denotes the probability that the first $i$ the labeling of the first sample. $y_i \in \{0, 1\}$; and $W_i$ denotes the category label of $y_i$ the coefficient of cost of misclassification of a sample with

$$W_i = \alpha_i \cdot e^{R_i} \quad (4)$$

Of these, the $\alpha_i$ and $R_i$ are fixed factors and adjustment factors after each iteration, respectively, which control the calculation of the loss function by $W_i$ controlling the calculation of the loss function and hence the model from following the distribution of the data more toward the majority sample.

Firstly, in equation (4) $\alpha_i$ represents the inverse of the number of samples in category i of the total number of samples, which is calculated as follows.

$$\alpha_i = \frac{\sum_{j=1}^{n} N_j}{c \cdot N_i} \quad (5)$$

where c is the hyperparameter $N_i$ denotes the number of the $i$th samples, and n denotes the total sample category. The sample is deterministic, so $\alpha_i$ is a fixed value, which is set by the degree of sample imbalance. In wheat frost detection, the sample imbalance is very serious (1:16), so we argue that the cost weight of the minority class samples always needs to be greater than the majority class weight, and $\alpha_i$ can make the cost weight of the minority class samples always greater than the cost weight of the majority class samples i.e. $W_1 > W_0$. Therefore, the model is always more concerned with wheat frost samples during the training process, which is consistent with the fact that in a real production environment one would be more concerned with samples suffering from disease damage.

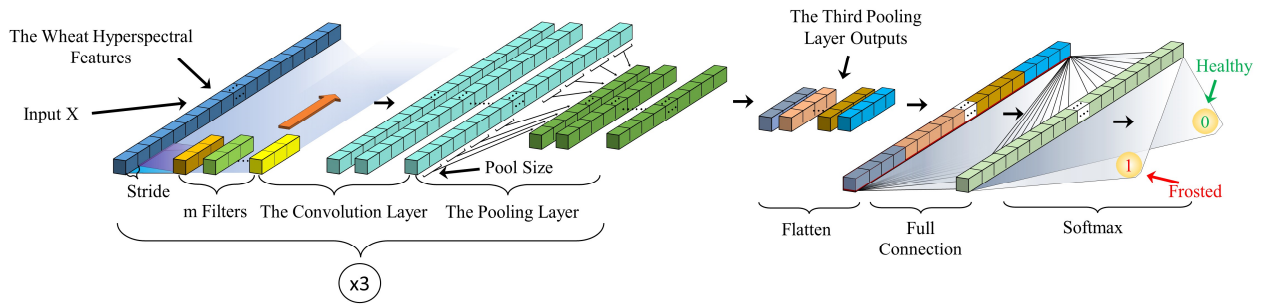

Figure 2. The 1D-CNN Model Architectur

Another parameter in equation (4) $R_i$ denotes the percentage of positive samples that were incorrectly predicted after each iteration and is calculated as

$$R_i = \begin{cases} \dfrac{N_{ij}}{N_{ii} + N_{ij}}, & i = 1 \\ 0, & i = 0 \end{cases} \quad (6)$$

where $N_{ij}$ denotes the number of samples in class $i$ that are predicted to class $j$. At i = 1, the initial value of $R_1$ is set to 1, i.e. it is assumed that the algorithm will not classify positive samples at the beginning, when $W_1$ maximum. In the subsequent training process, the probability of a positive sample being misclassified gets lower and lower, i.e. $R_1$ will become smaller and smaller, and the misclassification cost of positive samples will become lower and lower, eventually reaching an equilibrium state.

## III. EXPERIMENT AND RESULTS

### A. Experimental design

To verify the effectiveness of deep cost-sensitive learning in wheat frost detection, four comparative experimental schemes are designed in this paper: Scheme 1 uses a traditional machine learning algorithm to classify and detect the hyperspectral wheat frost dataset; Scheme 2 applies the cost-sensitive learning algorithm proposed in this paper to the hyperspectral wheat frost dataset; Scheme 3 removes the cost-sensitive learning from the model in Scheme 2 and uses it for the hyperspectral wheat. Scheme 4 uses a generative adversarial network to generate a certain number of samples to expand the hyperspectral wheat frost dataset, so that the number of frost samples and healthy samples in the adjusted dataset is approximately equal, and then uses a basic one-dimensional convolutional neural network to act on the new dataset to obtain experimental results. The specific experimental details of the above four Schemes are as follows.

Scheme 1: 70% of the samples in the hyperspectral wheat frost dataset are used as observation samples for traditional machine learning algorithms including SVM, KNN, and DT, and the remaining 30% of samples are tested. This division of the training set test set applies to all the schemes designed in this paper.

Scheme 2: In this paper, the PyTorch deep learning framework was used to implement the network in Fig. 2, and the GeForce RTX 3090 graphics card was used for training, with the total number of training sessions set to 1500. At the same time, to ensure equal input of the two types of samples, the smaller number of samples was replicated to ensure that the ratio of the number of samples for the two types of labels was kept within a certain range. In addition, this paper adopts Adam [17] as the gradient descent algorithm, with the learning rate initialized at 0.001 and decreasing to 10% of the current size after every 300 training sessions. A summary of the above parameters is shown in TABLE II.

TABLE II. TRAINING PARAMETER SETTING

| Parameter name | Parameter settings |
|---|---|
| epoch | $1.50 \times 10^3$ |
| batch_size | 256 |
| Learning rate | $1.0 \times 10^3$ |
| Learning rate decay | 0.100 |
| Activation function | ReLU |
| optimizer | Adam |
| c | 2.00 |

Scheme 3: This scheme is designed to verify the effectiveness of the cost-sensitive learning of Scheme 2. For this reason, cost-sensitive learning is deleted and all other settings remain the same as in scheme 1.

Scheme 4: This scheme is proposed to verify the validity of Scheme 2. A generative adversarial network is used as the basic model allowing samples to be added to adjust the data distribution. Specifically, given a random vector z (randomly sampled from a standard Gaussian distribution), z is fed into a neural network that outputs a feature (which can be a picture, text, etc.) The output of this scheme is a 2151 dimensional feature of

$$x^* = g(z) \quad (7)$$

where g(·) is the neural network that generates the samples; z denotes the input random vector, and $x^*$ denotes the generated sample, which are used as augmented data to supplement the original data for classification. The basic framework of this scheme is shown in Fig. 3. The generator and discriminator are trained with the dataset, then the expanded data generated by the generator is then combined with original data to form a new dataset for training the basic 1D-CNN network, and finally, the completed 1D-CNN model is used for frost detection on wheat.

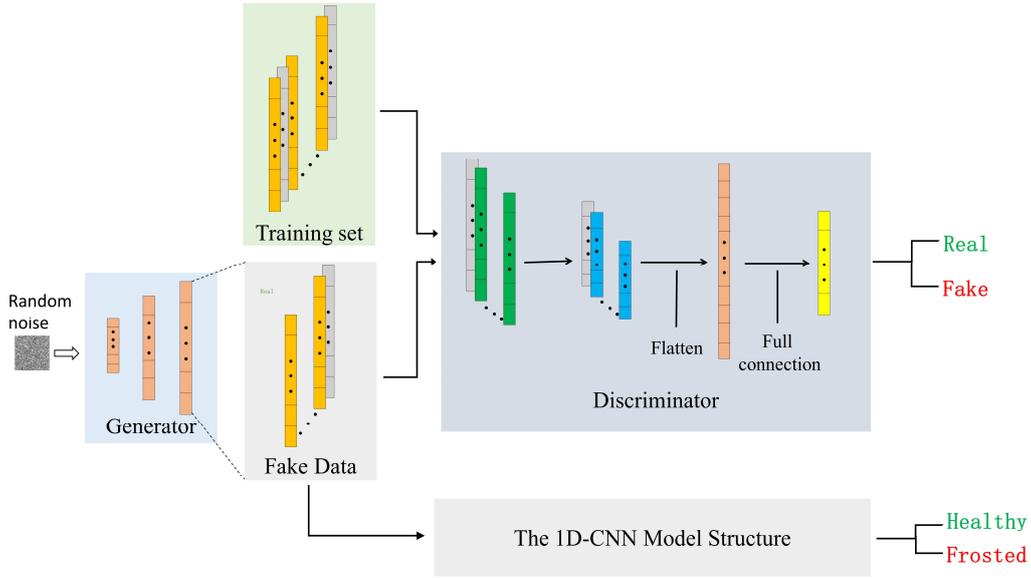

Figure 3. The Model Architecture of GANs

B. *Algorithm validation and analysis*

 1) *Evaluation indicators*

   *a) Accuracy:* The number of samples predicted correctly as a percentage of the total sample, is calculated as

$$Accuracy = \frac{1}{N}\sum_{i=1}^{m} N_i \quad (8)$$

where $N$ denotes the total number of samples, m denotes the number of sample label categories, and $N_i$ denotes the number of samples in category $i$ that were correctly predicted.

   *b) $F_1$-Score:* To take into account both precision and recall, we define a new metric: $F_1$-Score. $F_1$-Score takes into account both accuracy and recall, and the expression is

$$F_1 = \frac{2 * P * R}{P + R} \quad (9)$$

where $P$ denotes precision rate, $R$ denotes recall rate.

 2) *Experimental results and analysis*

   We ran the experiments corresponding to scenario 1, 2, 3 and 4 using the hyperspectral wheat frost dataset and the values obtained for each indicator are shown in TABLE III.

   As can be seen from the results in TABLE III, the traditional methods have high precision and accuracy rates, and the accuracy rates obtained using the SVM method even reach 100%, but its recall rates and $F_1$ scores are very low. This is because the traditional algorithms have a prerequisite assumption when classifying: the number of categories is approximately equal between them. This makes the traditional algorithms prone to misclassification of a small number of classes when faced with unbalanced data, so in the wheat frost detection task, the traditional algorithms cannot identify frost samples from a small number of classes, while the high dimensional nature of wheat hyperspectral data also adds to the classification difficulties. Combining the above descriptions, it emerges that the traditional algorithm in TABLE III has a low recall and $F_1$ scores are low, while accuracy and precision are high.

TABLE III. INDICATORS IN DIFFERENT METHODS

|  | Method | Accuracy/% | P/% | R/% | $F_1$/% |
|---|---|---|---|---|---|
| Traditional machine learning | SVM | 95.0 | 100 | 16.7 | 28.6 |
|  | KNN | 95.0 | 80.0 | 22.2 | 34.7 |
| Deep Learning | Baseline | 89.7 | 30.3 | 55.6 | 39.2 |
|  | GAN | 94.3 | **55.6** | 27.8 | 37.1 |
|  | CSBL(ours) | 94.3 | 51.9 | **77.8** | **62.3** |

   It can also be seen from TABLE III that applying deep learning can detect frosty samples better than traditional algorithms. At the same time, the cost-sensitive learning method was better in terms of accuracy and $F_1$ scores than other methods. Although the GAN method can provide more artificial samples for the classifier, the accuracy rate of the method was lower than the cost-sensitive learning method( $Accuracy$ =94.3%). This suggests that simply increasing the number of samples does not improve the confidence level of classification learning. On the contrary, the cost-sensitive learning method in this paper can improve the confidence of the classifier by adjusting the learning weights for different categories. In particular, when tested on the original model without cost loss, the accuracy rate also reached 94.3%, but its $F_1$ scores is low，indicating overfitting, in this case, the model improves accuracy rate by predicting all samples as healthy classes, but its recall $R$ and $F_1$ scores were significantly lower than those of this paper's method. Furthermore, the confusion matrix obtained using this paper's method (see TABLE IV) also demonstrates its ability to effectively classify a small minority of frost samples.

TABLE IV. CONFUSION MATRIX

| Baseline | | GAN | | CSBL (ours) | |
|---|---|---|---|---|---|
| 278 | 4 (0→1) | 259 | 23 (0→1) | 269 | 13 (0→1) |
| 13 (1→0) | 5 | 8 (1→0) | 10 | 4 (1→0) | 14 |

a. (0→1) indicates that a negative sample is predicted as a positive sample and (1→0) indicates that a positive sample. is predicted as a negative sample.

*3) Ablation study*

In order to investigate the effect of α and R on the overall effect of the depth cost-sensitive learning method in this paper, two more algorithms were designed separately

To investigate the effect of $\alpha_i$ and $R_i$ on the overall effect of the deep cost-sensitive learning method, we design two algorithms: (1) $\alpha_i$ remain unchanged and $R_i$ is set to 0; (2) $\alpha_i$ are set to 1 and $R_i$ unchanged. Each of these two algorithms was trained as a loss function for the basic convolutional neural network. It is easy to see that the first algorithm can be explored the role of $\alpha_i$ on the loss function, while the second algorithm can be designed in such a way as to investigate the effect of $R_i$ on the overall experimental results.

The experimental results obtained by applying these two algorithms to the hyperspectral wheat frost dataset are shown in TABLE V.

TABLE V. THE RESULT OF ABLATION EXPERIMENTS

| Methods | Accuracy/% | P/% | R/% | $F_1$/% |
|---|---|---|---|---|
| Baseline | 94.3 | **55.6** | 27.8 | 37.1 |
| +$\alpha_i$ | 92.7 | 40.0 | 66.7 | 50.1 |
| +$R_i$ | 93.0 | 42.7 | 44.4 | 43.5 |
| **CSBL(ours)** | **94.3** | 51.9 | **77.8** | **62.3** |

The experimental results in TABLE V demonstrate the importance of adding only $\alpha_i$ or $R_i$ the overall effect is improved compared to baseline, this is because $\alpha_i$ is a fixed factor that significantly improves the detection rate of frost samples by setting a higher misclassification cost for a small number of samples, thus increasing the lower limit of baseline. While the adjustment factor $R_i$ can adjust the misclassification cost of both in real-time according to the effect of each iteration, eventually reaching a balanced state, as shown in Fig. 4 $R_i$ changes dynamically during the training process, and at the initial training $R_i$ is large, and as training proceeds, the $R_i$ will reach a dynamic equilibrium within a certain range to achieve the best results. This paper combines the best of both worlds by using $\alpha_i$ The lower limit of the baseline is increased, and based on this, the overall results are adjusted using $R_i$ The overall results are adjusted so that the model is ultimately the most effective in detecting frosted wheat.

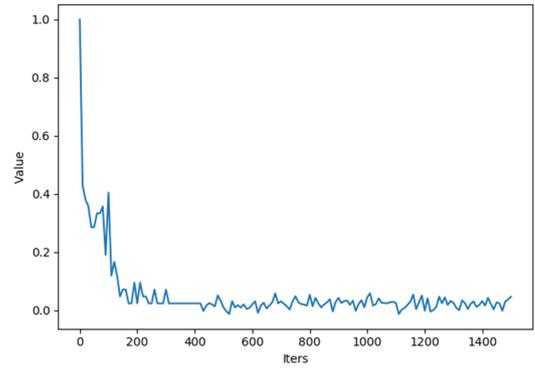

Figure 4. The value of $R_i$ during training

IV. CONCLUSION

In this paper, we construct a hyperspectral wheat frost dataset and propose a cost-sensitive learning method based on deep learning for the sample class imbalance problem in the data, with a one-dimensional convolutional neural network as the underlying framework, and by introducing cost coefficients into the loss function, the problem of model bias due to category imbalance during training is successfully overcome. At the same time, We also evaluate the model performance using a combination of accuracy and $F_1$. The results show that the deep cost-sensitive learning method in this paper can effectively identify frosty samples, and the experimental results are significantly better than the traditional machine learning algorithm, the basic 1D-CNN and GAN methods. It is an effective wheat frost detection algorithm.

ACKNOWLEDGMENT

This work was supported by Key Research and Technology Development Projects of Anhui Province (no. 202004a5020043). The financial support of academic support project for top-notch talents in disciplines (majors) of colleges and universities of Anhui Province (gxbjZD21081).